\documentclass[conference]{IEEEtran} 
\usepackage{cite}
\usepackage{amsmath,amssymb,amsfonts}
\usepackage{algorithmic}
\usepackage{graphicx}
\usepackage{textcomp}
\usepackage{xcolor}
\usepackage{soul}
\def\BibTeX{{\rm B\kern-.05em{\sc i\kern-.025em b}\kern-.08em
    T\kern-.1667em\lower.7ex\hbox{E}\kern-.125emX}}

\usepackage[utf8]{inputenc}
\usepackage{url}

\begin{document}

\title{A monitoring framework for deployed machine learning
models with supply chain examples}

\author{
\IEEEauthorblockN{Bradley Eck}
\IEEEauthorblockA{
\textit{IBM Research Europe}\\
Dublin, Ireland \\
bradley.eck@ie.ibm.com}
\and
\IEEEauthorblockN{Duygu Kabakci-Zorlu}
\IEEEauthorblockA{
\textit{IBM Research Europe}\\
Dublin, Ireland
}
\and
\IEEEauthorblockN{Yan Chen}
\IEEEauthorblockA{
\textit{IBM}\\
San Francisco, CA, USA
}
\and
\IEEEauthorblockN{France Savard}
\IEEEauthorblockA{
\textit{IBM}\\
Montreal, QC, Canada
}
\and
\IEEEauthorblockN{Xiaowei Bao}
\IEEEauthorblockA{
\textit{IBM}\\
Seattle, WA, USA
}

}

\IEEEpubid{978-1-6654-8045-1/22/\$31.00 \copyright 2022 IEEE}
\maketitle
\begin{abstract}
Actively monitoring machine learning models during production operations 
helps ensure prediction quality and detection and remediation 
of unexpected or undesired conditions.
Monitoring models already deployed in big data environments brings the additional challenges of adding monitoring in parallel to the existing modelling workflow
and controlling resource requirements. 
In this paper, we describe (1) a framework for monitoring machine learning models;
and, (2) its implementation for a big data supply chain application. 
We use our implementation to study drift in model features, predictions, and performance
on three real data sets.
We compare hypothesis test and information theoretic approaches to drift detection
in features and predictions
using the Kolmogorov-Smirnov distance and Bhattacharyya coefficient. 
Results showed that model performance was stable over the evaluation period.
Features and predictions showed statistically significant drifts; 
however, these drifts were not linked to changes in model performance during
the time of our study. 
\end{abstract}


\section{Introduction}

Monitoring of machine learning (ML) models in industrial applications helps ensure
prediction quality and integrity of business decisions made based on model predictions.  
Most models are trained under the assumption
of stationarity: that data used to make predictions will have the same probability 
distribution as that used for model training \cite{Ditzler15}.
However, industrial applications are often driven by non-stationary processes
due to seasonality, changes in consumer behavior, or evolving operational
conditions.  
High volume and velocity of data in these applications introduces additional challenges 
for model monitoring as the data of interest may only be available for a short time
or become expensive to access.  

Both the literature and commercial ML system developers recognize 
the need to monitor models during  production.
Synthesizing experience with numerous systems at Google,
Breck et al. weight monitoring during production as one quarter of their
overall score of model readiness \cite{Breck17}. 
Precisely which quantities to
monitor depends on the application. 
Considering the stationarity assumption underlying 
many models, it is not surprising that modern modelling software supports,
and recent studies investigate, several methods for detecting distribution shift.
Also called, drift detection, these methods evaluate the 
stationarity assumption by carrying out hypothesis tests 
such as the Kolmogorov-Smirnov test 
or estimating the shared information content using information entropy related metrics
such as the Kullback-Leibler divergence \cite{Huyen22}. Several studies present
comparisons of 
algorithmic approaches to drift detection for ML models 
 \cite{Rabanser2019} \cite{alibi-detect} \cite{Cobb22}.

The choice of which monitoring algorithms to apply for a given modelling task remains application dependant and so modelling
packages have growing support for drift related algorithms.
Tensorflow's data validation component quantifies drift using the L-infinity distance for categorical features and approximate Jensen-Shannon divergence for numeric features \cite{tensorflow2015-whitepaper}.
With pytorch, torchdrift supports several methods of  drift detection including the Kolmogorov-Smirnov and Max Mean Discrepancy tests \cite{torchdrift21}.
Scikit-learn (version 1.1.2) provides a variety of metrics to evaluate pairwise distances and sample affinity  \cite{sklearn11}.
Spark (version 3.3.1) provides the one-sample Kolmogorov-Smirnov test and several distance measures \cite{spark2016}.
The alibi-detect package provides algorithms for
outlier, adversarial and drift detection \cite{alibi-detect}.
With this landscape, much model monitoring can use algorithms from existing packages
but there is room to add methods tailored for big data use cases.

Several recent contributions examine  
strategies for implementing drift detection as part of the modelling life cycle.
Klaise et al. discuss the challenges of drift detection in production systems \cite{Klaise20}. 
The Augur framework \cite{Lewis22} examines drift detection metrics and thresholds with a view to eventually triggering model retraining.
The MLFlow \cite{mlflow18} tracking module, provides logging for metrics computed as part of a run.
The Castor time series forecasting system \cite{Chen2018} \cite{Eck2019} tracks performance of rolling predictions.
Drift monitoring also appears in the model lifecycle proposed by Hummer et al. who 
describe a cloud-based framework for AI Application development and lifecycle management \cite{Hummer19}.
In the market, cloud-based ML platforms provide monitoring for deployed models.
IBM's Watson\textregistered{}  OpenScale\texttrademark{} supports monitoring for bias, fairness, and drift \cite{openscale22}.  
Amazon Sagemaker\texttrademark{} provides a model monitor component that detects outliers and data drift \cite{sagemaker22}. 
Microsoft Azure\texttrademark{} also has drift detection for machine learning data sets \cite{azureML22}. 
At Google\textregistered{}, the Vertex AI model monitoring component handles drift detection for categorical and numerical features \cite{vertexai22}.
Although these platforms are feature rich, many existing software applications embed their machine 
learning workflows rather than use a cloud service. 
Moreover, existing commercial and open source model monitoring tools  
target models that already use a related software stack; there is thus gap for a more loosely coupled approach to model monitoring for applications with existing modelling workflows. 

Monitoring of production models thus requires both a framework suitable for the deployment
environment and metrics suitable for the modelling problem.  In the use cases that motivate
our work, we sought to add drift and performance monitoring in parallel to, and without disturbing, 
existing training and inference workflows.
On the algorithmic side, we sought metrics that could work from lightweight summaries of the data and use
computing and storage infrastructure already available in the target environment. 
We also sought to explore the application of hypothesis test and information theoretic metrics
for model monitoring in a supply chain use case
with a view to understanding
which metrics could anticipate changes in model performance.

This paper outlines two contributions to model monitoring.
First, we propose a framework for monitoring deployed models, especially where monitoring 
functionality should be added to applications that already embed model training and deployment. We motivate the framework with
applications from several domains using ML on big data.
In contrast to existing frameworks, we emphasize the deployment phase of the model
lifecycle. 
Second, we apply the framework to a supply chain use-case
and present  computational experiments on real world data.
These experiments use novel variations of classical techniques to
reduce computational and storage requirements for measuring
drift in features and predictions.  

The remainder of the paper is organized as follows. Section II outlines
the big data use-cases that motivate our work. Section III summarizes the
concepts and architectural design of our monitoring framework.  Section IV
describes our experiments leveraging big data tools including Spark, Parquet, 
and object storage to  monitor three models in a real supply chain use case.
Results results appear in Section V.
Finally, we conclude the work and note some promising directions for future effort in section VI.

\section{Motivating Applications}

Diverse applications, each with different modelling scenarios and
data types, motivate our work on monitoring ML models during production.
We consider monitoring models
that forecast sales in supply chains, predict failures of machinery, 
and classify objects on assembly lines. The following sections elaborate each application 
to inform requirements for model monitoring.

\subsection{Supply chain}

Sales forecasts help retailers order adequate quantities, analyze the effect of discounts
and position inventory to meet demand.
Sales of individual products are tracked by stock keeping unit (SKU) and location.  
Large retailers carrying many products at many locations have tens or
hundreds of millions of potential SKU-location combinations. 
Since every location will not carry every product the SKU-location mapping typically has low density. 
Table \ref{tab:datasets} summarizes three such real-world data sets. 

In our supply chain application, data scientists train new sales forecast models
each month using features as recent as the previous day's sales.
Each day, the inference workflow updates the model features and issues a new forecast. 
However, a model may start producing poor predictions
before the next scheduled training due to shifts in feature distributions,
buying habits, or other circumstances. 
For example, COVID-19 restrictions significantly changed consumer behavior.
Therefore,  these models require monitoring throughout deployment to enable model retraining
on a data-driven rather than scheduled basis.
Since a data scientist manages many supply chain forecasting models among several customers,
automation of monitoring across models is essential.

\begin{table}
    \caption{Summary of supply chain data sets considered for model monitoring.
             Dimension refers to the number of possible SKU-location combinations.
             Density refers to the percentage  of possible combinations 
             present in the data.}
    \label{tab:datasets}
    \centering
    \begin{tabular}{lrr}
    \hline
         Name & Dimension & Density \\
    \hline
         A  &   8,300,000 & 4\%   \\
         B  &  15,000,000 & 8\%   \\
         C  & 270,000,000 & 5\%   \\
    \hline
    \end{tabular}
\end{table}

\subsection{Equipment failure}

Industrial machines 
need maintenance to achieve optimal service life and return on investment.
The availability of sensor output from such equipment opens ML use cases 
such as failure prediction. 
In our application, sensors on the machines emit 
measurements which are then processed and used to predict asset health
and reliability. ML models making predictions with this sensor data can 
encounter non-stationarity for example when the unit of sensor output or 
operating regime changes. Hence the models require monitoring
especially to detect data drift.

\subsection{Object classification} 

In manufacturing, assembly lines produce products which require inspection 
for quality control and assurance. Image processing models doing object 
classification support human operators to deliver products with fewer defects.
In the application we study, training data sets for the model are typically small,
perhaps only a fraction of a percent of the number of images the model will
generate inferences on within its lifetime. Small variety in the training
data means that the image processing models receive 
myriad production images that are very different to training.
Model monitoring is thus crucial for delivering reliable predictions.

\section{Monitoring Framework}

Models for sales forecasting, equipment failure, and object classification
appear within existing enterprise software applications to provide 
automation and decision support. 
Synthesizing the steps needed to implement model monitoring,
the following steps were consistent across applications:
\begin{enumerate}
\item Register the model with the monitoring system and specify how the model should be monitored.
\item Process or store inference features, predictions, and optionally ground truth.
\item Compute metrics to evaluate model behavior and performance.
\item Evaluate metrics to trigger further actions like model retraining,
report generation, or alerting. 
\end{enumerate}

We describe the above steps with reference to the following concepts:
\begin{itemize}
\item \emph{Model}: the name or identifier for an instance of a trained ML
model that is monitored by the system.
\item \emph{Monitor}: a collection of metrics computed over the same data.
For example, a performance monitor could compare predictions with ground truth
using the mean average error, root mean square error, and other related metrics.
\item \emph{Metric}: a computed value that results from running a monitor. 
\item \emph{Reaction}: post-processing of metrics including side-effects.
For example, compare a metric to a threshold and send an alert.
\item \emph{Log}: a document that results from running a reaction
\end{itemize}

To enable use of the framework in applications with different data types and architectures,
we designed a layered system to separate the re-usable and application
specific logic. 
Highly re-usable components include the API that emerges from pairing 
the above concepts with verbs like get, set, run and delete and the key-value
schema for storing data generated during the production phase.

\subsection{System Design}

\begin{figure}
    \centering
    \includegraphics[width=3in]{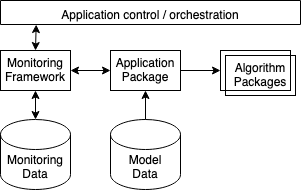}
    \caption{Architecture of monitoring framework.}
    \label{fig:arch}
\end{figure}

A loosely coupled ML model monitoring framework should be applicable to many domains
each with their own data types, computing environments, and storage infrastructure.
The framework proposed here aims to be flexible and extensible so that 
existing enterprise applications with ML models can easily add monitoring capability
with minimal disturbance to existing workflows.
The system comprises layers for
orchestration, monitoring logic, and data storage (Fig. \ref{fig:arch}).
The host application orchestrates monitoring of its ML models 
by invoking the monitoring framework. The framework in turn calls the 
application specific monitoring logic and 
stores the results in the monitoring data store. 
The application package provides
concrete implementations of the monitors and reactions needed for the target models.
These implementations typically use statistical hypothesis testing,
model evaluation, and dimensionality reduction algorithms provided by other libraries. 
To make these calculations, the application package also 
connects  to the model data storage.
The storage layer distinguishes data generated by monitoring system from data used to
train and score the model. 
Each of these components are further described in the following paragraphs.

The \textbf{framework package} encapsulates common functionality that interacts  with
the application specific package to deliver a working monitoring system.
Features of the framework package include the data schema, client interface, 
and abstract or base classes for the framework concepts of Monitors, Metrics, Reactions, and Logs. 
In our experience, these  features are independent of the ML model or application
specifications so that it is possible to reuse these components. 

The \textbf{application package} addresses functional needs of monitoring models in a
particular application. 
It provides concrete implementations of monitors, metrics, reactions and logs for the target modelling scenario.
For example, a report reaction enables custom visualisations for big data that plot samples of the data
instead of all points.
This package also handles the connection to the model training and inference data of the application.

\textbf{Algorithm package(s)} contain core implementations for computing metrics.  
Monitoring metrics depend on the data characteristics of an application.
For example, image data often requires dimension reduction while big data gets
most benefit from parallelism or approximated algorithms. Such algorithms are usually implemented without reference to the compute or storage infrastructure used by the data 
so that they can be used in multiple situations.

Finally, in the \textbf{data layer}, the framework package manages the storage
of monitoring data while the application package interacts with the model data.
For monitoring data,  
a key-value design allows storage of monitoring configuration and results using a
variety of storage technologies including IBM Cloud\textsuperscript{\textregistered{}} 
object storage, IBM DB2\textsuperscript{\textregistered}, MongoDB or a file system.
Model data, such as training data sets, features used to make predictions, values of predictions 
and eventually ground truth values remain in the model data storage. 
Applications with machine learning workflows already store this information; this design allows 
reuse of the existing data to support monitoring.

\subsection{Monitoring API}

A simple application programming interface emerges
from combining the framework concepts with the verbs 
set, get, run, delete. 
Setting a monitor associates  a monitor with a model. 
Running a monitor computes metrics for a particular model.
Getting metrics retrieves computed metrics from storage. 
Similarly,  setting a reaction associates a reaction and a model. 
Running the reaction creates logs. 
Logs may be the only result of running a reaction or may document a side-effect such
as sending an alert or triggering model retraining. 
Getting logs retrieves this information from the monitoring data storage.

Table \ref{tab:api} indicates the methods comprising our monitoring API. Most verb-object
combinations are supported. Exceptions are setting and running metrics and logs; these objects
result from running monitors and reactions and thus are not settable or runnable outside the framework.
With a layered system design and this API, the monitoring framework supports
the necessary workflow steps while allowing application-specific customization.

\begin{table}
    \caption{Monitoring API. 'x' indicates a supported command; empty cells mark unsupported commands.}
    \label{tab:api}
    \centering
    \begin{tabular}{lcccc}
    \hline
              & set & get & run & delete \\
    \hline
     Monitor  &  x  &  x  &  x  & x  \\
     Metrics  &     &  x  &     & x  \\
     Reaction &  x  &  x  &  x  & x  \\
     Logs     &     &  x  &     & x  \\
    \hline
    \end{tabular}
\end{table}

\section{Experimental Methodology}

We used the framework described above to carry out monitoring experiments
for the ML models making sales forecasts in our supply-chain use case.
These experiments had three related  aims.
First, we needed to quantify the performance of the sales forecast models  during production. 
Second, we wanted to check for non-stationarity in model features.
Third, we wanted to identify metrics that could be computed at forecast time
that might indicate an upcoming change in model performance. Our hypothesis was
that a statistical test for  distribution shift among the
features or predictions could be such an indicator.
To facilitate the experiments, we created a supply-chain application
package with monitors for drift detection and model performance.

\subsection{Drift Monitor} \label{subsec:driftmonitor}

We monitor features and predictions for distribution shift by comparing
data from training time to production values using variations of the 
Kolmogorov-Smirnov test and the Bhattacharyya coefficient. 
As further explained below, our variations to these classical methods enable
computation of these metrics in a distributed environment with Spark and 
generate summaries of the data for re-use  and visualization.

The Kolmogorov-Smirnov test for two samples calculates 
the largest distance between empirical cumulative distribution functions of the samples.
\begin{equation}
D_{KS} = max | F_1(x) - F_2(x) |
\label{eq:DKS}
\end{equation}
The distribution of this test statistic is also well known and so
p-values can be readily computed, for example using 
\cite{PressTeukolsky88}: 
\begin{equation}
  P(D_{KS}>observed) = Q_{KS}\bigg(\sqrt{\frac{N M}{N+M} D_{KS}}\bigg)
  \label{eq:PDks}
\end{equation}
where the quantile function is:
\begin{equation}
  Q_{KS}(\lambda) = 2 \sum_{k=1}^{\infty} (-1)^{k-1} e^{-2k^2 \lambda^2}
\end{equation}

In our variation of the test, we build an approximate cumulative distribution
function $\hat{F}$ for each sample. We construct $\hat{F}_1$ and $\hat{F}_2$
using approximate quantiles from the Greenwald-Khanna algorithm
\cite{Greenwald2001}  
as implemented in Spark's \texttt{approxQuantile} method.
We build $\hat{F}$ from quantiles at $a$ linearly spaced probabilities between $1/N$ and 1.
Evaluation at intermediate points uses linear interpolation between
the resulting quantiles. 
This approximation of $F$ summarizes the data of interest in fewer points.

The Bhattacharyya coefficient is defined by \cite{Kailath67} as 
\begin{equation}
    BC(p,q) = \int \sqrt{p(x)q(x)} dx,
    \label{eq:BC}
\end{equation}
where $p$ and $q$ are density functions. $BC$ represents the 
cosine of the angle between unit vectors representing distributions $p$
and $q$.
As a cosine, $BC=0$ indicates perpendicular unit vectors 
and hence probability distributions without overlap. Similarly $BC=1$ corresponds to 
parallel unit vectors and distributions that fully overlap.
Thus $BC$ is a convenient similarity measure for distributions as it always falls between 0 and 1, with 0 indicating no similarity and 1 indicating complete similarity.

In our variation, we evaluate $BC$ using estimated probability density functions, $\hat{f}$, derived
from the cumulative density estimates $\hat{F}$ already computed for (Eq. \ref{eq:DKS}).
The density estimate is
\begin{equation}
    \hat{f}(x) = (\Delta \hat{F}(x_k) + w_k) / \Delta x
\end{equation}
where $\Delta \hat{F}(x_k)$ is the difference in cumulative frequencies between
breaks of the $k^{th}$ bin;
$w_k$ is a small correction factor to ensure relative frequencies sum to 1; and 
$\Delta x$ is the bin width. In this way, the data summary $\hat{F}$ can be re-used to compute inputs for BC.

\subsection{Performance Monitor} \label{subsec:modelperfmonitor}

We monitor the in-production performance of sales forecasts from our model 
by comparing predicted and actual values.  The target variable of the forecasting
model is the seven-day average sales volume (termed velocity, as in units per day)
for a product at a location, $v_i$. 
Sales data become available each day and so after seven days the true value can be
computed.  

We report absolute and relative error metrics as follows.
The mean absolute error (MAE) is
\begin{equation}
    MAE =  \frac{1}{N} \sum_i^N \left|  \hat{v}_i - v_i \right|
    \label{eq:MAE}
\end{equation}
where the forecast velocity for the $i^{th}$ sku-location is $\hat{v}_i$; 
the true velocity is $v_i$; and there are $N$ sku-location combinations of interest.
Absolute errors are informative when target values have similar scales. When this is not the case, dividing the error by the target provides another useful view of performance.
When there are no sales of a product at a location for a week, the actual velocity $v_i$ is 0,
in which case the traditional MAPE cannot be computed. Thus we add a weight of unity to the true
value in the denominator and report a  weighted mean absolute percentage error:
\begin{equation}
    wMAPE = \frac{1}{N}  \sum_i^N \frac{\left|  \hat{v}_i - v_i \right|}{v_i+1} 
             \cdot 100\%
    \label{eq:wMAPE}
\end{equation}

Our workflow orchestrator runs the model performance monitor daily for deployed models.
Values of $\hat{v}_i$ are extracted from the model data store. Values of $v_i$ are computed on the fly 
from daily values in storage. We persist the resulting statistics for further analysis
and visualization.

\subsection{Run-time Environment}

\begin{figure*}
    \centering
    \includegraphics[width=6in]{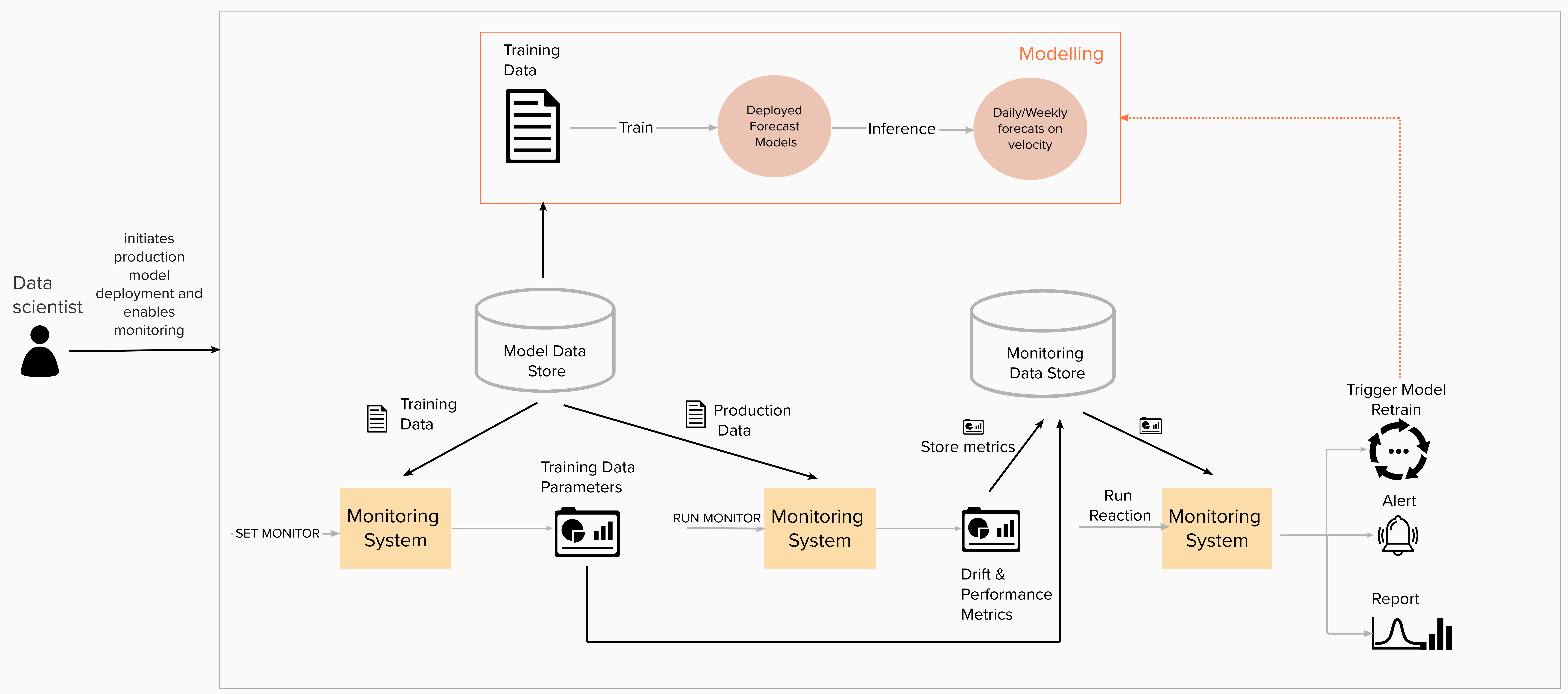}
    \caption{Modelling (top) and monitoring (bottom) workflows in our run-time environment.} 
    \label{fig:workflow}
\end{figure*}

Fig. \ref{fig:workflow} shows the parallel modelling and monitoring workflows in
our supply chain use-case.  
Data scientists initiate model training and deployment and enable monitoring.
For drift monitoring, the system evaluates training data and stores the approximate
cumulative distribution function, $\hat{F}$, of model features and predictions
along with other parameters in the monitoring data store for re-use.
Each day, new information becomes available and is used to make a new forecast; 
the monitoring system evaluates drift between training and production data. 
The resulting metrics are saved into the monitoring data store for later 
visualization and analysis.
Performance monitoring follows the same steps, except training data is not evaluated
because calculations of MAE and wMAPE need only predictions and ground truth values.

We run monitoring workflows on a compute cluster of 7 worker nodes each with 16 CPU cores and 58GB memory. Kubernetes version 1.22 allocates workload to compute nodes. We define and invoke model monitoring steps as Argo workflows.
Our algorithms are implemented in Python 3.7 using
pyspark\cite{spark2016}, numpy\cite{harris2020array}, scipy\cite{2020SciPy-NMeth}, pandas\cite{mckinney-proc-scipy-2010} and matplotlib\cite{Hunter:2007}. Separate buckets on IBM Cloud object storage contain
the model data, in Parquet\cite{Vohra2016} format, and monitoring data storage. 
Stocator\cite{stocator2018} provides the connection between Spark and object storage.  
Using the monitoring system in this runtime environment enables automation across model deployments.

\section{Results and Discussion}

\begin{figure}
    \centering
    \includegraphics[width=3.5in]{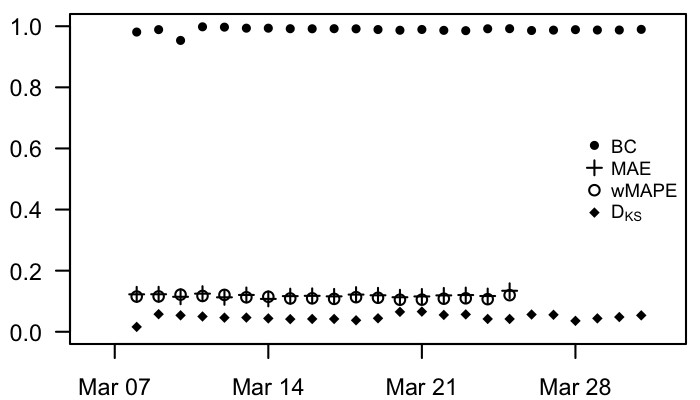}
    \caption{Prediction performance and drift metrics for data set A during March 2022.}
    \label{fig:metricsA}
\end{figure}

\begin{figure}
    \centering
    \includegraphics[width=3.5in]{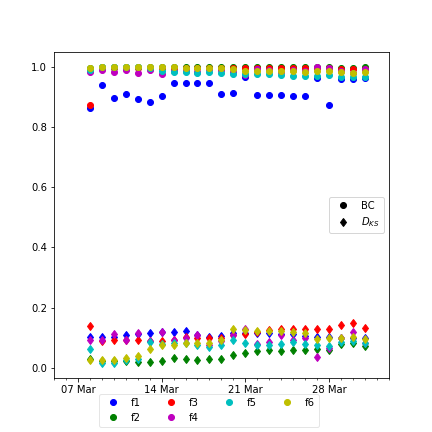}
    \caption{Drift metrics for six model features in data set A during March 2022.} 
    \label{fig:datadriftSetA}
\end{figure}

\begin{figure}
    \centering
    \includegraphics[width=3.5in]{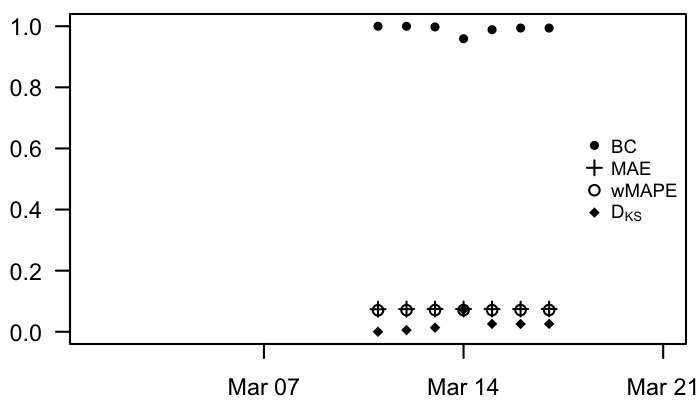}
    \caption{Prediction performance and drift metrics for data set B during March 2022.}
    \label{fig:metricsB}
\end{figure}

\begin{figure}
    \centering
    \includegraphics[width=3.5in]{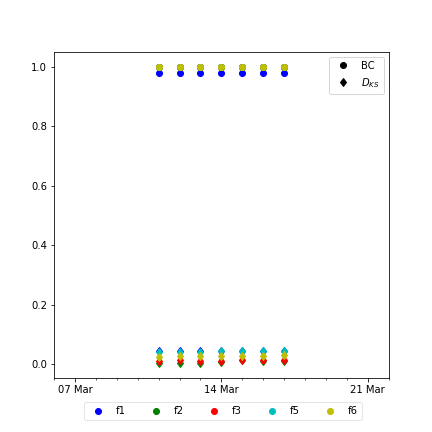}
    \caption{Drift metrics for five features of data set B during March 2022.} 
    \label{fig:datadriftSetB}
\end{figure}

\begin{figure}
    \centering
    \includegraphics[width=3.5in]{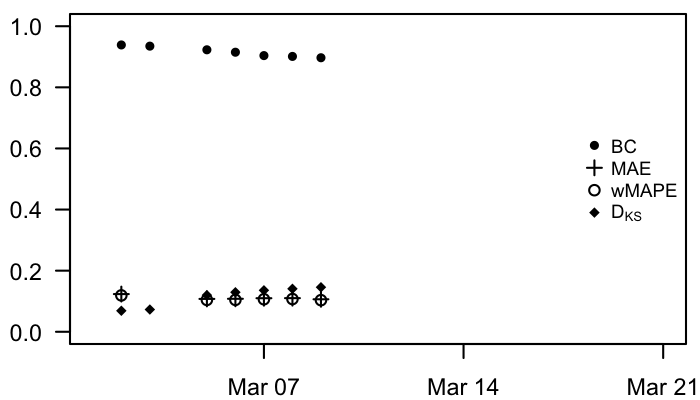}
    \caption{Prediction performance and drift metrics for data set C during March 2022.}
    \label{fig:metricsC}
\end{figure}

\begin{figure}
    \centering
    \includegraphics[width=3.5in]{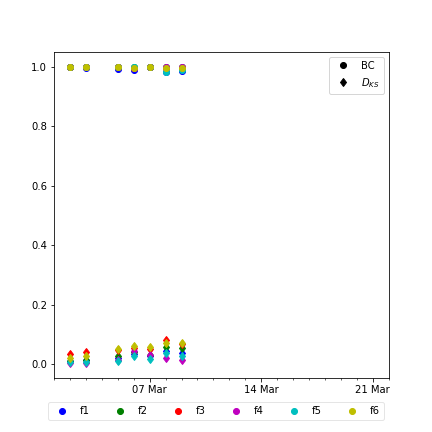}
    \caption{Drift metrics for six features of data set C during March 2022.}
    \label{fig:datadriftSetC}
\end{figure}

Our experiments aimed to quantify forecast model performance, check for non-stationary features, and explore 
metrics  that could be calculated at forecast time but
indicate an upcoming performance change.  
To address these questions we ran the drift and performance monitors outlined above on 
three supply chain examples. One model was trained for each data set. Six of the model
features are evaluated for drift.
Drift and performance metrics for data sets 
A, B, and C of Table \ref{tab:datasets}. 
during March 2022 appear in Figs. \ref{fig:metricsA} - \ref{fig:datadriftSetC}.
The figures show the Bhattacharyya coefficient, BC,  (Eq. \ref{eq:BC}) and
Kolmogorov-Smirnov distance $D_{KS}$ (Eq. \ref{eq:DKS}) for model predictions
and features by evaluation date.
For model performance, we plot MAE (Eq. \ref{eq:MAE})
and wMAPE (Eq. \ref{eq:wMAPE}) on the forecast date even though the computation was
actually carried out a week later, once the data became available. 
This arrangement compares the information available at forecast time with the 
model performance eventually observed.

We compare metrics' behavior across data sets using their coefficient of variation, $C_v$:
\begin{equation}
    C_v = \frac{s}{\Bar{x}}
\end{equation}
where the sample standard deviation is $s$ and the sample mean is $\Bar{x}$.
This ratio provides a convenient way of comparing dispersion between samples 
with different means.

For data set A, the model was trained with 3 months data on March 7th and used to make forecasts
 for the remainder of the month. 
Prediction performance (Fig. \ref{fig:metricsA}) was stable with MAE and wMAPE having similar coefficients 
of variation: 0.0483 and 0.047. The distribution of predictions was also quite 
stable, but the KS 
distance ($C_v=0.22$) showed considerably more variation than the Bhattacharyya coefficient
($C_v=0.0084)$. 
For model features (Fig \ref{fig:datadriftSetA}),
some trends of increasing $D_{KS}$ and decreasing $BC$  are visually evident in f2 and f6 while the other features showed little shift from the training distribution.
Values of $BC$ and $D_{KS}$ for model features showed more variation 
than predictions.  Neither drift in features or predictions were associated with 
changes in model performance.

For data set B, the model was trained with 3 months data on March 9th. Data was available for metric 
computation until March 17 except for March 10 which is not available.
Model performance (Fig. \ref{fig:metricsB}) was again stable with MAE and wMAPE showing similar coefficients
of variation 0.0065 and 0.0069 to each other, and slightly more variation than data set A.
For prediction drift,  KS distance ($C_v=0.98$) showed much more variation than  
the Bhattacharrya coefficient ($C_v=0.014$). The high variation in $D_{KS}$ was driven by
one very low value on March 11. March 14th showed highest shift in prediction values, but no change in model performance. Model features (Fig. \ref{fig:datadriftSetB}) did not show any drift; the March 14th movement in prediction values was not apparent in the features.  

For data set C, the model was trained with 1 month data on March 1st and production data available for the following week
except the 4th. A trend of increasing drift in predictions is visually evident in the declining BC
and rising $D_{KS}$ values (Fig. \ref{fig:metricsC}). Consistent with the other examples, that trend does not 
carry through to model performance, where MAE and wMAPE were stable and 
had similar  coefficients of 
variation of 0.057 and 0.055. KS distance ($C_v=0.28$) again showed more variation than
BC ($C_v=0.018$).  Drift metrics for model features (Fig. \ref{fig:datadriftSetC})
show a similar trend as those for predictions with drift most visually evident 
in f2 and f6.  Consistent with the other examples, variations in drift metrics of 
features and predictions were not associated with changes in model performance. 

Across data sets A, B, and C, model performance, as measured by MAE and wMAPE were 
very stable for the time periods over which data was available for these experiments.
While this result indicates models that perform well, the good model performance
hampered our experimental objective of identifying metrics that might anticipate 
changes in model performance.
The fact that we did not observe non-stationarity in model performance
may be due to the relatively short time period over which data were available. 
Some non-stationarity was evident in features of data sets A and C. However, the models coped
with this variation well and there were no meaningful changes in performance. 

\begin{figure}
    \centering
    \includegraphics[width=3.5in]{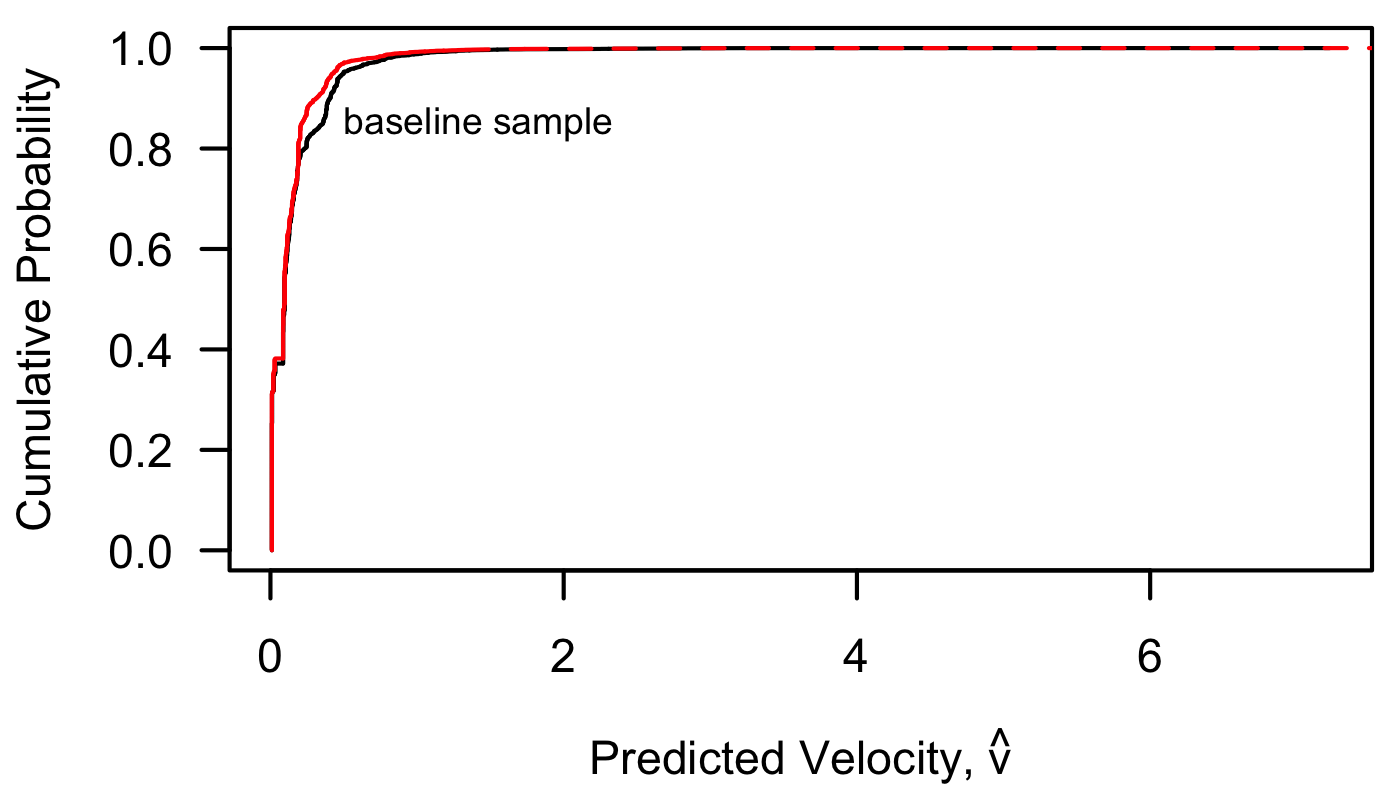}
    \caption{Cumulative distribution functions for predicted velocity in data set A on March 7 (training) and March 20 (production).}
    \label{fig:CDFs}
\end{figure}

Regarding the hypothesis that a statistical test for distribution shift could anticipate
changes in model performance, our experimental evidence did not support this view.
At the sample sizes considered here, even small values of $D_KS$  are highly significant. 
For example solving (Eq. \ref{eq:PDks}) with $\alpha=0.05$ and $N=M=332,000$ as for data set A
yields a critical distance  $D_{KS}=0.00333$. Similarly, using $N=M=13,500,000$ as for data set C
gives a distance of $D_{KS}=0.0005227$.  
An example is instructive to see why a highly significant distribution shift may not effect
model performance. 
Figure \ref{fig:CDFs} shows cumulative distribution functions for model predictions in the 
training  and production
samples from data set A on March 20, where the KS distance is 0.065. This distance is highly statistically 
significant with a p-value < $10^{-16}$. While this distance is visible
on the plot, in this context it is understandable  that the distribution has not changed in an operationally meaningful way. For this reason, we also report BC values to provide evidence that the shape of the distribution has not changed much. 

\section{Conclusions}

We presented and applied a framework for monitoring machine learning
models during deployment to three supply chain examples.
The framework enables adding model monitoring capability
to an existing application that is already training
and scoring of ML models. It uses the application's storage infrastructure and supports calculation of metrics tailored to the use case.

In our supply chain examples we analyze sales predictions and model
features for distribution shift. 
Across three data sets in March 2022, the forecast model performance
as measured by MAE and wMAPE was very stable.  The distribution of 
predicted sales showed more variation than model performance but
this variation did not translate into performance changes,
suggesting that the models are performing as intended. 
Features showed more drift but this was also not connected with performance change.
All 
of the KS tests reported here were highly statistically significant.
The fact no tests were operationally meaningful
sounds a cautionary note about the utility of hypothesis testing 
for drift detection in this application. 
As an additional metric, the Bhattacharyya coefficient provided useful confirmation
that the distribution shape had not changed much, despite the KS distance.

Future work could apply the proposed monitoring framework to other applications
and longer duration data sets to identify suitable thresholds for BC, KS or other metrics to support 
alerting and retraining.

\bibliographystyle{IEEEtran}

\bibliography{mmbd}

\end{document}